\documentclass{pprai}
\usepackage[utf8]{inputenc}
\usepackage[T1]{fontenc}
\usepackage{graphicx}
\usepackage{amsthm}
\usepackage{txfonts}
\usepackage{url}
\usepackage{algpseudocode}
\usepackage{booktabs}

\title{Cartesian Genetic Programming Approach for Designing Convolutional Neural Network}
\headtitle{Cartesian Genetic Programming Approach for Designing CNN.}
\author{Maciej Krzywda$^{1}$, Szymon {\L}ukasik$^{12}$ and Amir H. Gandomi$^{34}$}
\headauthor{M. Krzywda, S. Lukasik, A. H. Gandomi}
\affiliation{
  $^1$Faculty of Physics and Applied Computer Science, AGH University of Krakow, al. Mickiewicza 30, 30-059 Krak\'{o}w, Poland\\
  $^2$Systems Research Institute, Polish Academy of Sciences, ul. Newelska 6, 01-447 Warsaw, Poland\\
  $^3$ Faculty of Engineering and IT, University of Technology Sydney, 5 Broadway, Ultimo NSW 2007, Australia\\
  $^4$University Research and Innovation Center (EKIK), Óbuda University, Bécsi út 96/B, Budapest, 1034, Hungary
}
\begin{document}
\maketitle
\begin{abstract}
The present study covers an approach to neural architecture search (NAS) using Cartesian genetic programming (CGP) for the design and optimization of Convolutional Neural Networks (CNNs).  In designing artificial neural networks, one crucial aspect of the innovative approach is suggesting a novel neural architecture. Currently used architectures have mostly been developed manually by human experts, which is a time-consuming and error-prone process. In this work, we use pure Genetic Programming Approach to design CNNs, which employs only one genetic operation, i.e., mutation. In the course of preliminary experiments, our methodology yields promising results.
\end{abstract}

\section{Introduction}
Neural Architecture Search (NAS)\cite{Elsken2018NeuralAS} has gained considerable traction as a fully automated approach in the design of Neural Networks Architecture. The method facilitates the generation of architectures that are not only comparable but often superior in performance to those crafted manually. Essentially, NAS simplifies the traditional process where humans iteratively adjust neural networks through trial and error to identify successful configurations. Instead, it automates this process, unveiling more intricate structures. Comprising a spectrum of techniques and tools, NAS systematically evaluates numerous network architectures within a predefined search space. It employs a search strategy to choose the architecture that best fulfills the objectives of a specific problem, maximizing a fitness function. Despite its effectiveness, NAS poses significant computational and time-related challenges, exacerbated by the financial costs associated with utilizing graphics processing units (GPUs). Consequently, researchers and research groups are increasingly exploring alternative methods to optimize costs and identify the most efficient and effective neural network architecture tailored to address their specific research problems.
This paper aims to investigate and present results for designing convolutional neural networks using Cartesian Genetic Programming (CGP).

\section{Cartesian Genetic Programming for Convolutional Neural Network Design}
Pure Cartesian Genetic Programming does not include the crossover operation, so the only genetic operation in CGP is a mutation.  
Using crossover is a hot topic of research to improve GCP, but is still widely investigated \cite{10.1007/978-3-319-77553-1_13}. Several types of mutations can be applied in CGP, such as point mutation, gene mutation, and segment mutation. Point mutation involves randomly changing the value of a single gene in the genome, whereas gene mutation involves replacing an entire gene with a new one. Segment mutation involves replacing a genome segment with a new segment that may contain multiple genes\cite{Miller2019CartesianGP,Miller2011}. Here, we present the use of Cartesian Genetic Programming (CGP) for designing convolutional neural networks (CNNs). Our goal was to establish the architecture of the first neural network so that it could be compiled. Still, we did not define it according to existing solutions - we wanted the network design process to establish solution on its own. In our approach, the evolution process starts with a parent whose genotype is represented by a set of genes and information about which layers in the neural network are active. Then, two offsprings are created by mutating the parent's genotype, where mutations are performed with a specific mutation rate. Children resulting from the mutation process are judged to see whether they are better than their parents, and if so, that child is the parent of the next generation. We have predefined ConvSets with usable layers, each of which is decoded. The randomly selected layer is replaced with another random layer from ConvSets. If a random number is generated that is the same as the original value (the layer selected to replace cannot be replaced by the same one), the function will generate a new random value. We then check that this change allows compilation and that all shapes between layers are correct. If so, the mutation process is complete. If the mutation results s not accepted, and we draw another set of genes to mutate. Then, each offspring is converted to a neural network, and their performance is compared with the parent. If any offspring achieves better performance than the parent, it is considered the new parent, and its genotype is used to generate the next generation of offspring. Otherwise, the parent remains unchanged. The comparison of mutations and performance is repeated for several generations. In each generation, the best genotype found is recorded and the value of the fitness function represents the best performance. If none of the offspring achieves better performance than the parent for several consecutive generations, a neutral mutation is applied to the parent and the evolution process continues.  However, to keep the layers active, we need to check that the mutation will not affect the deactivation of the layers, which is essential for the correct operation of the neural network. If the mutation results in deactivating these layers, the mutation is not accepted, and we draw another set of genes to mutate. Then, each offspring is converted to a neural network, and their performance is compared with the parent. If any offspring achieves better performance than the parent, it is considered the new parent, and its genotype is used to generate the next generation of offspring. Otherwise, the parent remains unchanged. The comparison of mutations and performance is repeated for several generations. In each generation, the best genotype found is recorded and the value of the fitness function represents the best performance. If none of the offspring achieves better performance than the parent for several consecutive generations, a neutral mutation is applied to the parent and the evolution process continues. Our goal was and is still to explore potential spaces where we can discover the optimal network configuration, leveraging the evolution process along with the mutation process. This allows the neural network genotype to be optimized and aproaches us the best model for a given dataset. Fitness is defined as a measure of how well a particular neural network performs on a given task. In this example, the fitness function is the accuracy of the network in a validation dataset. The higher the accuracy, the higher the fitness of the network. During the evolution process, the goal is to maximize fitness by generating new candidate solutions (children) through mutation and selecting the best ones for the next generation based on their fitness compared to their parents.

\section{Experiments}

\subsection{Experimental Setting and Computational Budget}
In our research, we employed various activation functions, including Relu, Elu, Selu, Sigmoid, Softmax, Softplus, Softsign, Tanh, and Exponential. Additionally, we utilized the following pooling methods: GlobalAveragePooling2D, MaxPool2D, and AveragePooling2D. We applied a dropout rate of 0.2 and incorporated BatchNormalization with a momentum of 0.99 and epsilon of 0.001. The parameters of the GCP algorithm utilized in our experiments include rows set to 1, Cols equal to 30, Level-Back set to 10, and both Mutation rate and generation specified as values of the list [0.01, 0.05, 0.1] and [10, 25, 50] respectively.

\begin{equation}
    Budget= P \times G \times E \times T
    \label{eq:budget}
\end{equation}

To compare our solution, we define the computational budget as follows (equation \ref{eq:budget}) where the population (P)  was the count of chromosomes representing a single CNN. The evaluation epoch (E) is a count of the epoch to train the mutated CNN in each generation (G) iteration. Training epochs (T) track the best chromosome trained during the epoch after the last iterate of generation. We set the population as one. We define our budget, where in our experiments E (Evaluation Epochs) is 25. Evaluation Epochs are used in the evolution process when we try to decode CNN architectures that can be compiled and all shapes between layers are correct; then we train our architecture in 25 epochs, on a tiny part of the data (1000 records) (in a 70/30 ratio) and verify fitness. Then we compare the achieved fitness value with others achieved in a given generation, and the best individual (with the lowest fitness) is trained in 100 epochs (T).

In our approach, we use 10, 25, or 50 generations (G) and always 100 epochs (T) to train the best convolutional neural network. To avoid the overfitting effect, we implement an Early Stopping mechanism for monitoring loss function (Categorial Cross Entropy) with patience set as 10. Each of the best neural network architectures was trained 10 times and the final results were averaged and presented in Table \ref{tab:results}. Data were shuffled each time in 80/20 proportion. Therefore, the possible computational budgets are as follows:

\begin{itemize}
    \item For G=10: Budget = 1 * 10 * 25 * 100 = 25000 = \textbf{25K}
    \item For G=25: Budget = 1 * 25 * 25 * 100 = 62500 = \textbf{62.5K}
    \item For G=50: Budget = 1 * 50 * 25 * 100 = 125000 = \textbf{125K}
\end{itemize}

\subsection{Experimental results}
\label{sec:results}
\begin{table}[htb]
  \centering
    \caption{Cartesian Genetic Programming parameters}
  \begin{tabular}{|c|c|c|c|c|c|}
    \hline
    \textbf{Parameter} & Rows & Cols & Level-Back & Mutation rate & Generation \\ \hline
    \textbf{Value} & 1 & 30 & 10 & [0.01, 0.05, 0.1] & [10,25,50] \\ \hline
  \end{tabular}
  \label{tab:CGPparams}
\end{table}

\begin{figure}[htb]
  \centering
  \includegraphics[width=1.in]{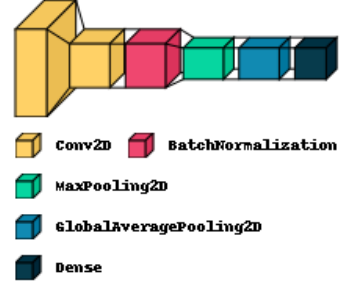}
  \includegraphics[width=1.in]{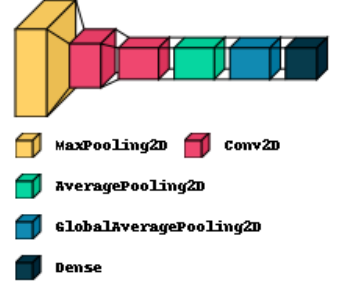}
  \includegraphics[width=1.in]{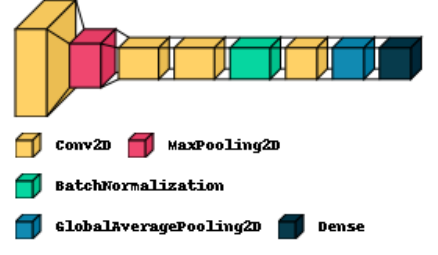} \\
  \includegraphics[width=1.in]{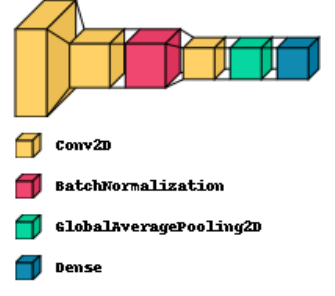}
  \includegraphics[width=1.in]{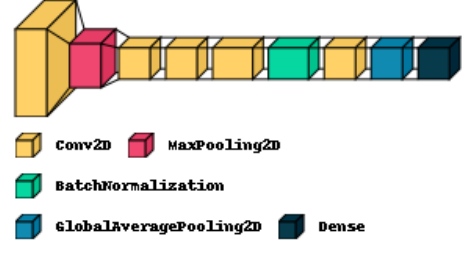}
  \includegraphics[width=1.in]{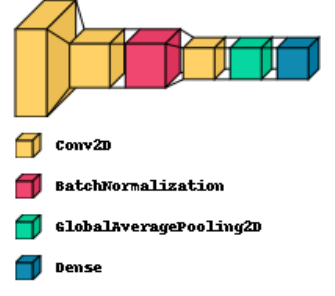}
  \caption{Visual representation of a stack of layers in Designed Convolutional Neural Network with CGP}
  \label{fig:networks}
\end{figure}

To validate the robustness of the results, an experiment was conducted in which the parameters listed in Table \ref{tab:CGPparams} were used, and the experiment was repeated ten times. The results of this experiment as CNNs stacked with layer recognition are presented in (Table \ref{fig:networks}). In the table presented (Table \ref{tab:results}) we demonstrate our result compared to CNN-based selected baseline solutions using classification images from the MNIST, and Fashion-MNIST datasets depending on the defined budget of the proposal and depends on mutation rate (MR). We also present all the results obtained, depending on the defined budget and mutation ratio. We can see that the medium budget \textbf{62.5K} yields the best results, while for a budget of 125K, the algorithm performs the worst. The number of generations determines the budget size; therefore, you can conclude that the increase in generations influences the degradation of the results. Based on the experimental settings and results, we can define the same budget using a larger population of neural networks instead of increasing generations. But at this stage of research, it is only a reckless hypotesis, which needs more evidence including more run experiments with a different range of parameters, including a smoother range of Mutation rates and bigger population, not only one.

\begin{table}[ht]
\centering
\caption{Accuracy in \% achieved by our models}
\begin{tabular}{lcc}
\toprule
\textbf{Models} 						& \textbf{MNIST} & \textbf{FashionMNIST}  \\
\hline 
  Our CNN (\textbf{25K},MR=\textbf{0.1}) 		&92.51 $\mp$ 0.351		& 86.87 $\mp$ 0.201  \\
  Our CNN (\textbf{62.5K},MR=\textbf{0.1}) 	&97.28 $\mp$  0.842	& 72.82 $\mp$  0.435\\
  Our CNN (\textbf{125K},MR=\textbf{0.1}) 	&55.29 $\mp$ 1.113		& 44.82 $\mp$  1.384\\ \hline
  Our CNN (\textbf{25K},MR=\textbf{0.01}) 	&97.15 $\mp$ 0.373		& 83.27 $\mp$  0.883\\
  Our CNN (\textbf{62.5K},MR=\textbf{0.01}) 	&97.75 $\mp$ 0.220		& 70.95 $\mp$  0.538\\
  Our CNN (\textbf{125K},MR=\textbf{0.01}) 	&54.43 $\mp$ 0.984		& 86.44 $\mp$  0.271 \\ \hline
  Our CNN (\textbf{25K},MR=\textbf{0.05}) 	&97.92 $\mp$ 0.119      & 83.73 $\mp$  0.339 \\
  Our CNN (\textbf{62.5K},MR=\textbf{0.05}) 	&92.25 $\mp$ 0.237	    & 72.10 $\mp$  0.471\\
  Our CNN (\textbf{125K},MR=\textbf{0.05}) 	&91.82 $\mp$ 0.325 		& 84.06 $\mp$  0.792\\ \hline
  L.Hertel et al. (2017)\cite{7280683}				      & 99.68 & N/A \\
  L.Wan et al. (2013) \cite{Wan2013RegularizationON}		  & 99.79 & N/A \\
  K.Meshkini et al. (2020) \cite{9047776}				      & N/A & 93.43 \\
  M.Kayed et al. (2020) \cite{10.1007/978-3-030-50097-9_10} & N/A & 98.80 \\
\end{tabular}
\label{tab:results}
\end{table}

\section{Conclusions}

In the last few years, studies on designing Artificial Neural Networks (ANNs) have become an active research field, mainly due to the advanced cost training and prototyping of underlying deep learning architectures.
Our approach, distinguished from more complex solutions \cite{Torabi2022UsingCG,Suganuma2020EvolutionOD,Banerjee2020EvolvingOC}, focuses on simplicity. We employ basic CNN layers, disregard data dimensionality post flattening, and incorporate residual connections. Additionally, our method seamlessly integrates state-of-the-art models as input architecture, facilitating the analysis of changes in future generations.The proposed study focuses on developing the first stage of a method for the design and optimization of artificial neural networks based on Cartesian Genetic Programming. The focus is on developing an efficient and effective approach to designing and optimizing ANNs.
\section*{Acknowledgment}
This work was partially supported by the program 'Excellence Initiative - Research University' for the AGH University of Krakow and by a Grant for Statutory Activity from the Faculty of Physics and Applied Computer Science of the AGH University of Krakow. We gratefully acknowledge the Polish high-performance computing infrastructure PLGrid (HPC Centers: ACK Cyfronet AGH) for providing computer facilities and support within computational grant no. PLG/2022/015677"

\bibliography{pprai}

\begin{thebibliography}{10}
\providecommand{\url}[1]{\texttt{#1}}
\providecommand{\urlprefix}{URL }
\expandafter\ifx\csname urlstyle\endcsname\relax
  \providecommand{\doi}[1]{doi:\discretionary{}{}{}#1}\else
  \providecommand{\doi}{doi:\discretionary{}{}{}\begingroup \urlstyle{rm}\Url}\fi

\bibitem{Elsken2018NeuralAS}
Elsken, T., Metzen, J.~H., and Hutter, F.
\newblock Neural architecture search: A survey.
\newblock \emph{ArXiv}, abs/1808.05377, 2018.

\bibitem{10.1007/978-3-319-77553-1_13}
Husa, J. and Kalkreuth, R.
\newblock A comparative study on crossover in cartesian genetic programming.
\newblock In M.~Castelli, L.~Sekanina, M.~Zhang, S.~Cagnoni, and P.~Garc{\'i}a-S{\'a}nchez, editors, \emph{Genetic Programming}, pages 203--219. Springer International Publishing, Cham, 2018.
\newblock ISBN 978-3-319-77553-1.

\bibitem{Miller2019CartesianGP}
Miller, J.~F.
\newblock Cartesian genetic programming: its status and future.
\newblock \emph{Genetic Programming and Evolvable Machines}, 21:129--168, 2019.

\bibitem{Miller2011}
Miller, J.~F.
\newblock \emph{Cartesian Genetic Programming}, pages 17--34.
\newblock Springer Berlin Heidelberg, Berlin, Heidelberg, 2011.
\newblock ISBN 978-3-642-17310-3.
\newblock \doi{10.1007/978-3-642-17310-3_2}.

\bibitem{7280683}
Hertel, L., Barth, E., Käster, T., and Martinetz, T.
\newblock Deep convolutional neural networks as generic feature extractors.
\newblock In \emph{2015 International Joint Conference on Neural Networks (IJCNN)}, pages 1--4. 2015.
\newblock \doi{10.1109/IJCNN.2015.7280683}.

\bibitem{Wan2013RegularizationON}
Wan, L., Zeiler, M.~D., Zhang, S., LeCun, Y., and Fergus, R.
\newblock Regularization of neural networks using dropconnect.
\newblock In \emph{International Conference on Machine Learning}. 2013.

\bibitem{9047776}
Kayed, M., Anter, A., and Mohamed, H.
\newblock Classification of garments from fashion mnist dataset using cnn lenet-5 architecture.
\newblock In \emph{2020 International Conference on Innovative Trends in Communication and Computer Engineering (ITCE)}, pages 238--243. 2020.
\newblock \doi{10.1109/ITCE48509.2020.9047776}.

\bibitem{10.1007/978-3-030-50097-9_10}
Meshkini, K., Platos, J., and Ghassemain, H.
\newblock An analysis of convolutional neural network for fashion images classification (fashion-mnist).
\newblock In S.~Kovalev, V.~Tarassov, V.~Snasel, and A.~Sukhanov, editors, \emph{Proceedings of the Fourth International Scientific Conference ``Intelligent Information Technologies for Industry'' (IITI'19)}, pages 85--95. Springer International Publishing, Cham, 2020.
\newblock ISBN 978-3-030-50097-9.

\bibitem{Torabi2022UsingCG}
Torabi, A.~R., Sharifi, A., and Teshnehlab, M.
\newblock Using cartesian genetic programming approach with new crossover technique to design convolutional neural networks.
\newblock \emph{Neural Processing Letters}, 55:5451--5471, 2022.

\bibitem{Suganuma2020EvolutionOD}
Suganuma, M., Kobayashi, M., Shirakawa, S., and Nagao, T.
\newblock Evolution of deep convolutional neural networks using cartesian genetic programming.
\newblock \emph{Evolutionary Computation}, 28:141--163, 2020.

\bibitem{Banerjee2020EvolvingOC}
Banerjee, S. and Mitra, S.
\newblock Evolving optimal convolutional neural networks.
\newblock \emph{2020 IEEE Symposium Series on Computational Intelligence (SSCI)}, pages 2677--2683, 2020.

\end{thebibliography}
\bibliographystyle{pprai}

\end{document}